\documentclass{article}
\usepackage{ijcai19}

\usepackage{times}
\usepackage{soul}
\usepackage{url}
\usepackage[hidelinks]{hyperref}
\usepackage[utf8]{inputenc}
\usepackage[small]{caption}
\usepackage{graphicx}
\usepackage{amsmath}
\usepackage{booktabs}
\usepackage{algorithm}
\usepackage{algorithmic}
\title{Mappa Mundi: An Interactive Artistic Mind Map Generator with Artificial Imagination}

\author{
Ruixue Liu$^1$\and 
Baoyang Chen$^2$\and
Meng Chen$^1$\and
Youzheng Wu$^1$\and
Zhijie Qiu$^2$\And
Xiaodong He$^1$
\affiliations
$^1$JD AI, Beijing, China\\
$^2$Central Academy of Fine Arts, Beijing, China
\emails
\{liuruixue,chenmeng20,wuyouzheng1,xiaodong.he\}@jd.com, \{chenbaoyang, qiuzhijie\}@cafa.edu.cn
}

\begin{document}

\maketitle

\begin{abstract}
We present a novel real-time, collaborative, and interactive AI painting system, Mappa Mundi, for artistic Mind Map creation. The system consists of a voice-based input interface, an automatic topic expansion module, and an image projection module. The key innovation is to inject Artificial Imagination into painting creation by considering lexical and phonological similarities of language, learning and inheriting artist's original painting style, and applying the principles of Dadaism and impossibility of improvisation. Our system indicates that AI and artist can collaborate seamlessly to create imaginative artistic painting and Mappa Mundi has been applied in art exhibition in UCCA, Beijing\footnote[1]{http://ucca.org.cn/en/exhibition/qiu-zhijie-mappa-mundi/}.
\end{abstract}
\section{Introduction}
Mind Map is a diagram that connects relevant words, terms or events around a keyword or an idea and it's perceived as an artistic possibility representation of knowledge. Through constant interpretation and abstraction, the artistic creation of Mind Map is a process of subject decision making and information expansion. In this paper, we introduce Mappa Mundi, an interactive Mind Map generator featured with artificial imagination.

The creation of Mind Map requires the interaction between AI and artist on the decision making of idea expansion. While previous AI image generators, such as pix2pix \cite{c100} framework, depends only on the model performance and less on taking artist's interaction into account. Besides, models constructed with neural approaches barely extract features from big data, whereas features, such as imagination and creativity, are far to reach. As Mind Map represents the universal mind of human thinking, the Kantian thinking of free play of imagination, which is the foundation of modern arts \cite{c99}, plays an important role in expanding artist's thought and connecting the flourishing information. Different from the traditional framework, Mappa Mundi offers opportunities to enable AI and artist together during art making. 

Mappa Mundi enables voice input from human and extracts the keywords for vivid expansion on myriad levels of domains, which explores the impossibility of improvisation variation arising from standardisation. 
To improve the imaginative feature of word expansion results, Mappa Mundi employs not only semantic similarity and linguistic features but also establishes a knowledge graph to imitate artist's idea and style. Besides, Mappa Mundi breaks conventions and explores more possibilities for word's connections partially stem from the Dadaism art movement. Through interaction sparked in our system, the free play of human imagination and machine intelligence is consolidated, and eventually constitutes a contemporary Mappa Mundi, which cultivates a more and more electrified arena for endless imagination of both human and machine.

\section{Related Work and Challenges}
Nowadays, AI demonstrates stronger potential for art creation. Many researches have been conducted to involve AI into poem generation \cite{c20,c21}, creation of classical or pop music \cite{c27,c28} and automatic images generation \cite{c24,c25,c30}. Whereas there are  few researches exploring the possibility of artificial imagination for artwork creation. 

To generate an imaginative and creative Mind Map, the key problem is automatic topic expansion. There are several challenges: first of all, language is an informative system. Thus many linguistic features should be considered in words and relation extension. Second, as an artwork, it should reflect artist's mind and individual experience. Third, for Mind Map creation, artists usually pay less attention on defining accurate relation between concepts, instead, they are defined with imagination. So, to build artistic Mind Map, the system should find a way to break the convention rules for the connection between literal representations.

\section{System Architecture}
\begin{figure}
  \centering
  \includegraphics[width=1\linewidth]{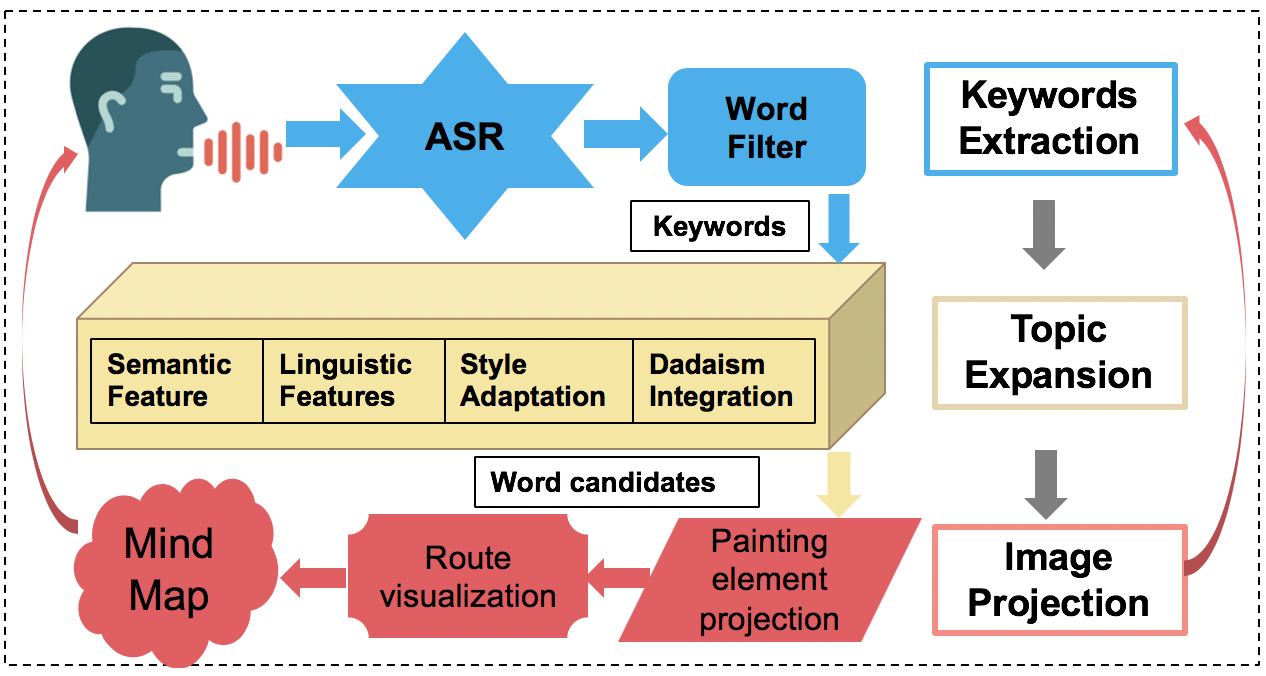}\\
  \caption{System architecture of Mappa Mundi}
   \setlength{\abovecaptionskip}{0.cm}
   \setlength{\belowcaptionskip}{-0.cm}
  \label{fig:system}
\end{figure}

The overall architecture of Mappa Mundi is illustrated in Figure \ref{fig:system}. It consists of three modules in total. The first one is a voice-based input interface, which translates user's voice input into text and extracts keyword from it. The second is topic expansion module, which expands the keyword into several word candidates that share various connections with it. The third module is image projection, which takes the keyword and its candidates as input and projects the concepts and relations into a Mind Map image. As an interactive system, Mappa Mundi presents artist's speech as Mind Map image and this imaginative picture in return provokes artist's aspiration for further thoughts expansion.  

\subsection{Voice-based Input Interface} Automatic Speech Recognition (ASR) engine is designed for interaction. 
Both English and Chinese languages are supported. We also performed lexicon and language model adaption to enhance the recognition of words in art domain. After that, we developed a keyword extraction function to extract meaningful words from the converted text for further expansion. 
We adopted open-source software jieba\footnote[2]{https://github.com/fxsjy/jieba} for Chinese and Stanford parser \cite{stanford} for English POS tagging. Only informative words such as noun, verb and adjective words are kept and TFIDF weights are calculated for further filtering. The output of this module is the keywords.

\subsection{Topic Expansion}
As imagination is the soul for artistic Mind Map, Mappa Mundi employs several features to increase information variety \cite{c0} during topic expansion. It firstly uses word embeddings to find candidates based on semantic similarity \cite{c8,c9,c10}. To enrich linguistic information of expansions, it also takes the morphological and phonological features into account. Words sharing similar characters or morphemes or phonetic syllables are selected as candidates. Moreover, 
we construct a knowledge graph (KG) by extracting concepts and relations from artist's original Mind Map masterpieces. Mappa Mundi will adopt those concepts and relations once a topic word is covered by the KG. By doing so, it can better imitate artists' mind and thoughts. Further, Mappa Mundi breaks the restriction of domains and conventions and explores more possibilities for words' connections by creating rules following the famous Dadaism principle in art movement \cite{dada}. Finally, with all above mentioned methods, Mappa Mundi can create a branch of creative and informative word candidates for a given keyword. 

\subsection{Image Projection} 
To visualize the abstract concepts, Mappa Mundi includes around 3000 painting elements which are extracted from the representative Mind Map paintings of a famous artist\footnote[3]{https://en.wikipedia.org/wiki/Qiu\_Zhijie}. 
These elements are sharing traditional Chinese Shan-shui (Mountain-river) painting style and are further analyzed into 5 typical Shan-shui painting categories including \textit{architecture, mountain, river, grassland and lake}. 
Mappa Mundi will classify each word into one of the five categories above and then connect it with the elements in the corresponding domain. An additional type of painting element \textit{Route} is also involved. The distance between the concept and it's candidate is determined by their similarity score and then visualized as \textit{Route} in image. The more similar these two words are, the closer distance will be. 
\begin{figure}
  \centering
  \includegraphics[width=\linewidth]{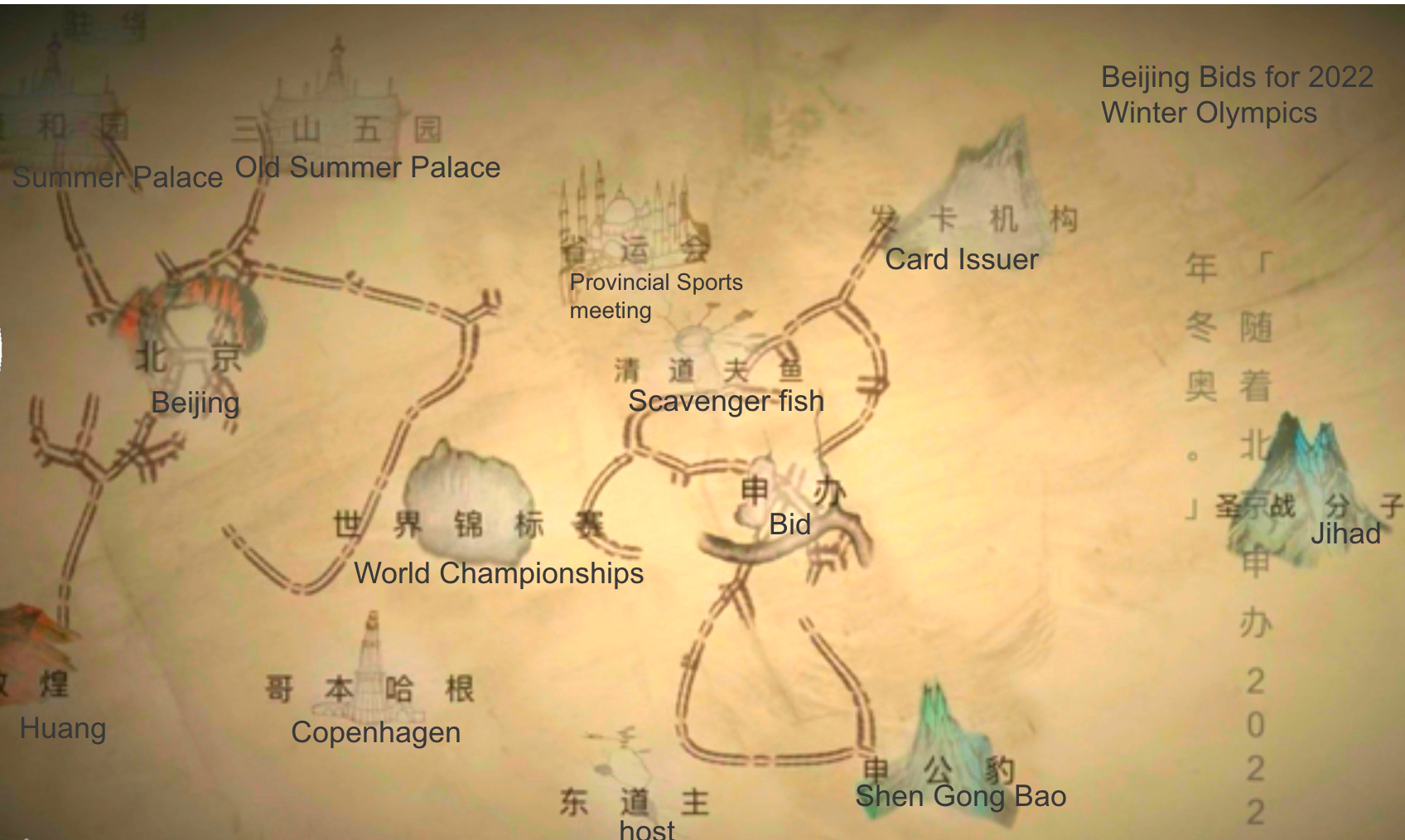}\\
  \caption{Screen-shot of a generated Mind Map image with voice input of \textit{Beijing Bids for 2022 Winter Olympics}}
   \setlength{\abovecaptionskip}{0.cm}
   \setlength{\belowcaptionskip}{-0.cm}
  \label{fig:demo1}
\end{figure}

Our Mappa Mundi is not a one-shot completion, it is an interaction of live stimulations, which contains constant vocal inputs via ASR and outputs of generated words and their relations. It is a two-way street. The generated information, after being presented in our system, becomes the inspiration for artist's next vocal input. This artwork reflects both the development of artist's thinking and the AI-enabled imagination. 
\section{Conclusion}
In this paper, we propose a Mind Map generator, Mappa Mundi, that can inject Artificial Imagination together with human interaction into an artwork. The way of enriching artificial imagination includes considering semantic similarity, integrating informative linguistic features, inheriting artist's mind, and inserting Dadaism and impossibility of improvisation principles. Mappa Mundi is presented to understand how AI can aid artistic practice. Its development also indicates AI has more implications in art than a way of artistic creation, and it enriches the possibility of artistic practice.
\section*{Acknowledgments}
We want to thank Yan Dai, Chuyan Xu, Haozhen Feng and Xiaoyu Guo for their work and support to build this system.  

\bibliographystyle{named}
\bibliography{ijcai19}

\end{document}